\documentclass[final]{cvpr}

\usepackage{times}
\usepackage{epsfig}
\usepackage{graphicx}
\usepackage{amsmath}
\usepackage{amssymb}
\usepackage[accsupp]{axessibility}
\usepackage{ifpdf}
\usepackage{url}
\usepackage{algorithm}
\usepackage{algorithmic}
\usepackage{textcomp}
\usepackage{xcolor}
\usepackage{subfigure}
\usepackage{siunitx}
\usepackage{booktabs}
\usepackage{multirow}
\usepackage{float}
\usepackage{caption}
\usepackage{threeparttable}
\usepackage{diagbox}
\usepackage{bm}

\usepackage[pagebackref=true,breaklinks=true,colorlinks,bookmarks=false]{hyperref}

\def\cvprPaperID{35} 

\def\cvprPaperID{35} 
\def\confYear{CVPR 2022}

\begin{document}

\title{SwinIQA: Learned Swin Distance for Compressed Image Quality Assessment}  

\author{Jianzhao Liu, Xin Li,Yanding Peng, Tao Yu, Zhibo Chen \\
\textit{CAS Key Laboratory of Technology in Geo-spatial Information Processing and Application System
}\\
\textit{University of Science and Technology of China}\\
{ \small \tt \{jianzhao, lixin666, pyd, yutao666\}@mail.ustc.edu.cn, chenzhibo@ustc.edu.cn}}

\maketitle
\begin{abstract}
   Image compression has raised widespread interest recently due to its significant importance for multimedia storage and transmission. Meanwhile, a reliable image quality assessment (IQA) for compressed images can not only help to verify the performance of various compression algorithms but also help to guide the compression optimization in turn.  In this paper, we design a full-reference image quality assessment metric SwinIQA to measure the perceptual quality of compressed images in a learned Swin distance space.   
   It is known that the compression artifacts are usually non-uniformly distributed with diverse distortion types and degrees. To warp the compressed images into the shared representation space while maintaining the complex distortion information, we extract the hierarchical feature representations from each stage of the Swin Transformer. 
   Besides, we utilize cross attention operation to map the extracted feature representations into a learned Swin distance space. Experimental results show that the proposed metric achieves higher consistency with human's perceptual judgment compared with both traditional methods and learning-based methods on CLIC datasets.
\end{abstract}
\vspace{-0.5cm}
\section{Introduction}
Image/Video compression plays a pivotal role in modern society. Currently, there are various compression methods including traditional codecs (\textit{e.g.}, HEVC/H.265 \cite{hevc}, VVC/H.266 \cite{vvc}) and learning-based methods~\cite{wu2020learned,gao2021perceptual}, which aim to solve rate-distortion optimization (RDO) problem. In such a process, IQA of compressed images plays a vital role in guiding the optimization and verification of various compression algorithms.
\begin{figure}[htbp]
 \centering
 \subfigtopskip=1pt
 \subfigbottomskip=-0.2pt
 \subfigure[]{
  \label{fig:ref}
  \includegraphics[width = .14\textwidth]{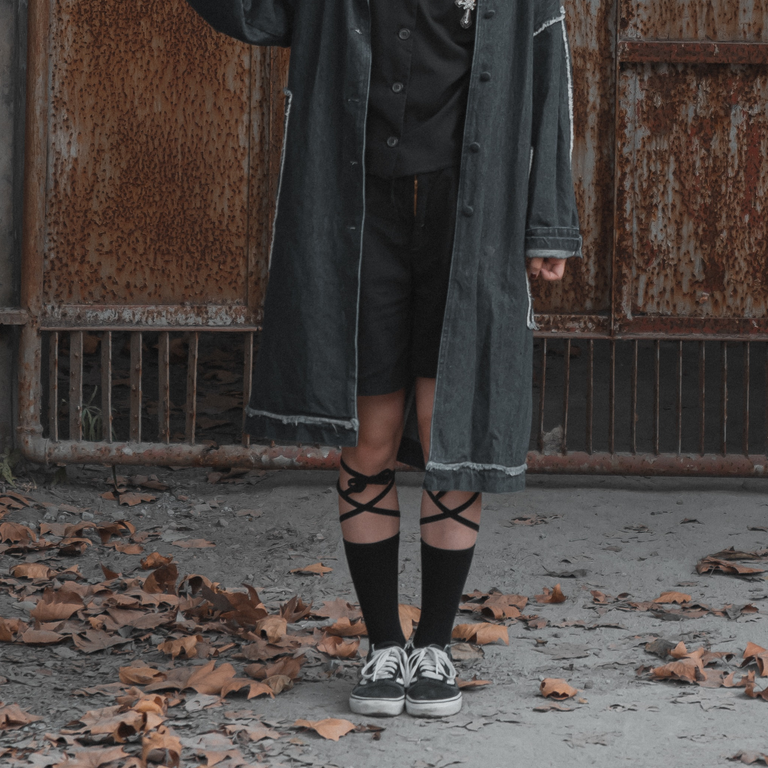}}
 \subfigure[]{
  \label{fig:dist1}
  \includegraphics[width = .14\textwidth]{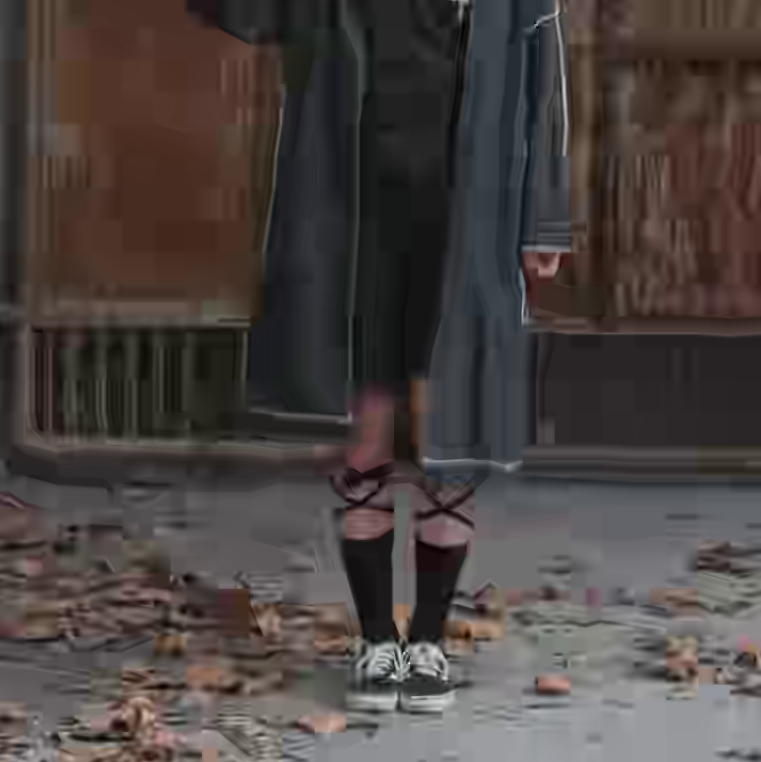}}
  \subfigure[]{
  \label{fig:dist2}
  \includegraphics[width = .14\textwidth]{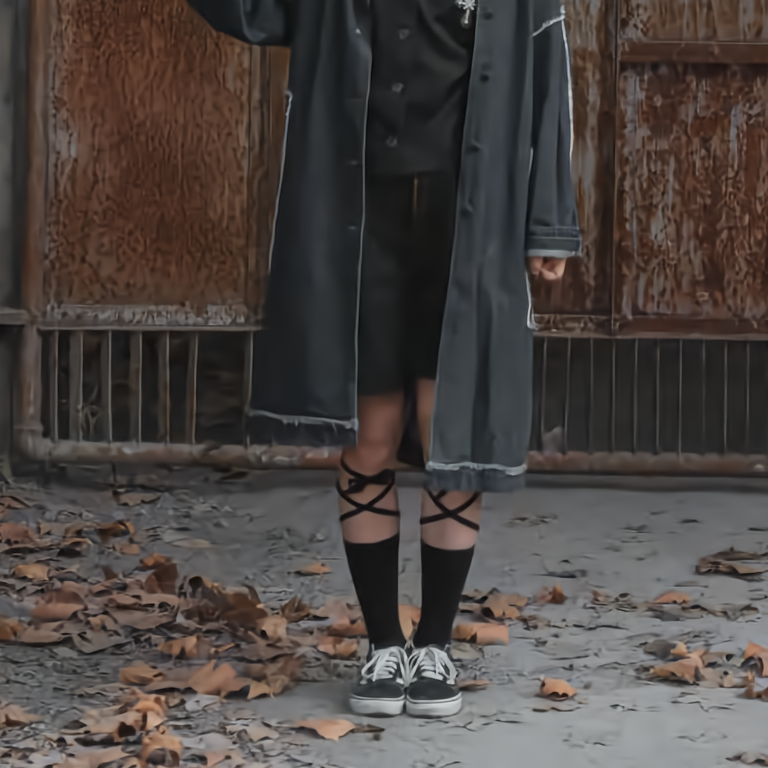}}
\caption{Illustration of the compression artifacts. (a) Reference image. (b) Distorted image  generated by HEVC codec. (c) Distorted image  generated by learning-based codec.  }
\label{fig:hevc}
\vspace{-0.3cm}
\end{figure}

Commonly used traditional IQA algorithms in image compression methods, such as PSNR (peak signal-to-noise ratio), are mainly utilized to measure the pixel-wise fidelity. Though they have low computational complexity, they are not well matched to perceived visual quality.
Structure similarity (SSIM) index~\cite{ssim} measures the patch similarity between the reference and the distorted images, based on the assumption that the human visual system (HVS) tends to perceive the local structures. It achieves more consistent results with human perceptual quality on popular datasets. 
Moreover, learning-based metrics also show impressive improvement\cite{liu2021liqa,peng2021multi}. LPIPS~\cite{LPIPS} obtains the perceptual similarity judgment by calculating the ${l}_{2}$ distance between features extracted from deep convolutional neural networks (CNNs) pre-trained on ImageNet classification task. Similarly, DISTS~\cite{DISTS} measures the texture and structure similarities between the VGG-based deep features to calculate the perceptual similarity of two images, which achieves the state-of-the-art (SOTA) performance on benchmark datasets. 

Recently, Transformer has shown promising potential in computer vision area and outperforms CNN in various mainstream tasks such as image classification and object detection. Taking advantage of the self-attention layer, Transformer can capture long-range pixel interactions and aggregate the global information from the entire input sequence.  Vision Transformer (ViT) \cite{ViT} splits an image into patches and treats the image patches as tokens (words) to input to a Transformer following the same way in an NLP application. However, the complexity of ViT can increase quadratically with the number of image patches. 
To tackle this challenge, Swin Transformer \cite{liu2021swin} is designed by integrating the advantages of both CNNs and Transformers. By limiting self-attention computation to non-overlapping local windows, it has the advantages as CNN to process images with large size due to local attention mechanism. By allowing for cross-window connection it has the advantages as Transformer to model the long-range dependencies in the data.

Inspired by the success of Transformer, several researchers attempted to apply transformers in the IQA task. TRIQ\cite{you2021transformer} utilizes
 a shallow Transformer encoder on the top of a feature map extracted by CNN for blind image quality assessment. IQT \cite{cheon2021perceptual} extracts the feature representations from a CNN backbone and then feds the extracted feature maps into the transformer encoder and decoder in order to compare the reference and the distorted images. As shown in Fig.~\ref{fig:hevc}, the compression artifacts are usually non-uniformly distributed with diverse types and degrees, thus it is important to combine the local-global information to measure the perceptual quality of compressed images. In this paper, we propose a full-reference image quality assessment metric named SwinIQA, based on Swin Transformer. We demonstrate that the hierarchical features extracted from each stage of the Swin Transformer have strong representation ability towards the non-uniformly distributed compression artifacts. Besides, instead of calculating the $l_{2}$ distance or feature similarity between the reference and the distortion image features like LPIPS or DISTS, we utilize cross attention to map the extracted feature representations into a learned Swin distance space.
Experiment results show that our SwinIQA achieves state-of-the-art performances on CLIC2022 validation set and CLIC2021test-subtest. Moreover, we also conduct experiments of different distance mapping strategies to verify the effectiveness of the cross attention operation when comparing the reference and the distorted features.

\begin{figure*}
    \centering
	\includegraphics[width=0.95\textwidth]{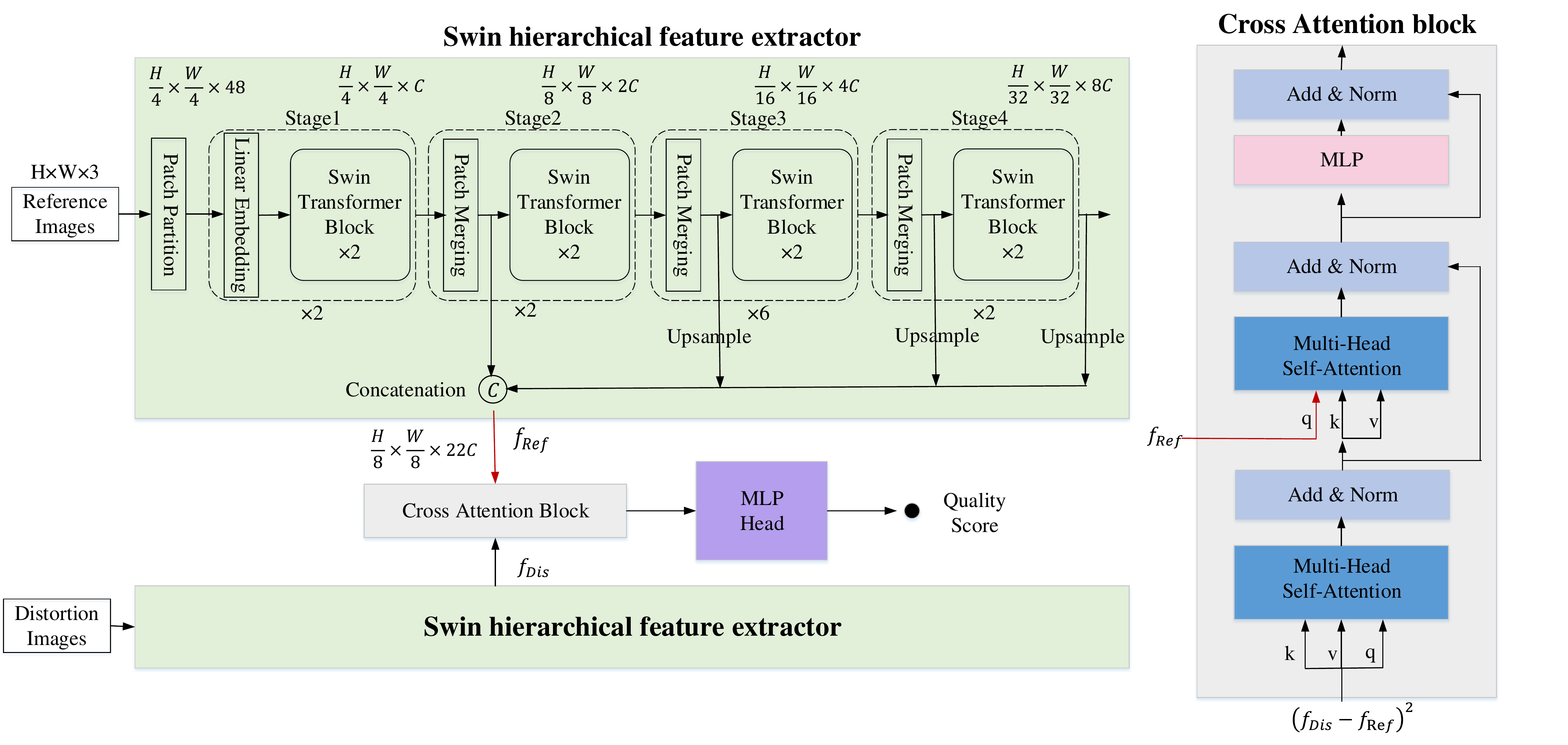}
    \caption{Framework of SwinIQA. Zooming in for better viewing.}
    \label{fig:network arch}
    \vspace{-3mm}
\end{figure*}

\section{Approach}
In this section, we will introduce the architecture of our SwinIQA first. Then we will introduce the training strategy of our method. 

\subsection{Network Architecture}
The framework of SwinIQA is shown in Fig.~\ref{fig:network arch}. It consists of three parts: a Swin hierarchical feature extractor that extracts multi-scale local-global feature representations, a cross attention block that maps the pair of reference and distortion feature representations into a learned Swin distance space, and a MLP head which maps the learned Swin distance into a quality score.

As shown in Fig.~\ref{fig:network arch}, Swin Transformer builds hierarchical
feature maps by merging multi-level deep features. Considering that the compression artifacts are usually non-uniformly distributed with diverse distortion types and degrees, we utilize Swin Transformer as the feature extractor to extract the multi-scale hierarchical representations. Given an image $I\in \mathbb{R}^{H\times W\times 3}$, we first extract intermediate features from each stage of the Swin Transformer and obtain a group of features
$\{f_{1}\in \mathbb{R}^{\frac{H}{8}\times \frac{W}{8}\times 2C},f_{2}\in \mathbb{R}^{\frac{H}{16}\times \frac{W}{16}\times 4C},f_{3}\in \mathbb{R}^{\frac{H}{32}\times \frac{W}{32}\times 8C},f_{4}\in \mathbb{R}^{\frac{H}{32}\times \frac{W}{32}\times 8C}\}$. 
Then we upsample all the features to  ${\frac{H}{8}\times \frac{W}{8}\times 2C}$ and concatenate the features along the channel dimension to get the final hierarchical feature representations $f\in \mathbb{R}^{\frac{H}{8}\times \frac{W}{8}\times 22C}$ :
\begin{equation}
f=[{{f}_{1}},Up({{f}_{2}}),Up({{f}_{3}}),Up({{f}_{4}})],
\end{equation}
where $[\ ]$ denotes the concatenation operation, $Up$ means upsampling operation, \textit{e.g.} bilinear upsampling.

For full-reference IQA task, given a reference image $I_{Ref}$ and a distorted image $I_{Dist}$,  their hierarchical feature representations are denoted as $f_{Ref}$ and $f_{Dist}$, respectively. In order to better measure the perceptual distance of $f_{Ref}$ and $f_{Dist}$, we adopt cross attention operation to map the feature representations of the reference image and the distortion image to a learned Swin distance space. The cross attention operation is defined by:
\begin{equation}
\label{equ:cross attention}
\begin{aligned}
  & {z}'=LN(MHSA(q,k,v)) \\ 
 & q={{{({{f}_{Dis}}-{{f}_{\operatorname{Re}f}})}^{2}}}{{W}_{q}},k={{{({{f}_{Dis}}-{{f}_{\operatorname{Re}f}})}^{2}}}{{W}_{k}},\\&v={{{({{f}_{Dis}}-{{f}_{\operatorname{Re}f}})}^{2}}}{{W}_{v}}, \\ 
 & {z}''=LN(MHSA({q}',{k}',{v}')+{z}'), \\ 
 & {q}'={{f}_{\operatorname{Re}f}}{{{{W}'}}_{q}},{k}'={z}'{{{{W}'}}_{q}},{v}'={z}'{{{{W}'}}_{v}}, \\ 
 & {{f}_{mapped}}=LN(MLP({z}'')+{z}''), \\ 
\end{aligned}
\end{equation}
where $LN$ represents LayerNorm, $MHSA$ represents the standard multi-head self-attention module in a transformer. $MLP$ consists of several Fully-connected layers. $q$, $k$ and $v$ denote the query, key and value respectively. It should be noted that in the second $MHSA$ module of the cross attention block, we use the reference feature $f_{Ref}$ as the query. 
Finally, a MLP regression head is employed to regress the  ${{f}_{mapped}}$ in the learned Swin distance space to a perceptual quality score:
\begin{equation}
d = MLP({{f}_{mapped}})
\end{equation}

\subsection{Training Strategy}
We first pretrain the SwinIQA on the KADID-10K \cite{KADID-10K} dataset, which contains MOS value for each of the distorted images.  We adopt MSE loss for training:
\begin{equation}
\label{equ:reg}
    L_{reg}={{\left\| D({{I}_{\operatorname{Ref}}},{{I}_{Dist}})-s \right\|}_{2}},
\end{equation}
where $D$ denotes the proposed SwinIQA which compares the perceptual distance of the image pair ${I}_{\operatorname{Ref}}$ and ${{I}_{Dist}}$. $s$ denotes the ground-truth normalized MOS value ($s=1.0-MOS/5.0$ for KADID-10K). Higher $s$ denotes larger perceptual distance and worse perceptual quality compared with the reference image.

Then we recruit datasets which employ two alternative forced choice (2AFC) test. It means that these datasets only contain labels describig which of two distorted images is more similar to a reference.
Given a triplet $(I_{Ref},I_{Dist1},I_{Dist2})$, we should compute $d_{1}=D(I_{Ref},I_{Dist1})$ and $d_{2}=D(I_{Ref},I_{Dist2})$ to decide which image is of higher fidelity compared with the reference image. Following the work of LPIPS \cite{LPIPS}, given two distances $d_{1}$ and $d_{2}$, we utilize a small judgment network $\mathcal{G}$ to map the distance feature $[d_{1},d_{2},d_{1}{-}d_{2},d_{1}/d_{2},d_{2}/d_{1}]$ to a  predicted judgment score $\hat h \in (0,1)$. The architecture uses two 32-channel $FC-ReLU$ layers followed by a 1-channel $FC$ layer and a sigmoid function. We adopt Binary Cross Entropy (BCE) loss for training:
\begin{equation}
\label{equ:bce}
\begin{aligned}
    L_{bce}(I_{Ref},I_{Dist1},I_{Dist2},h)=\\-h\log \mathcal{G}(D(I_{Ref},I_{Dist1}),&D(I_{Ref},I_{Dist2}))\\
    -(1-h)\log (1-\mathcal{G}(D(I_{Ref},&I_{Dist1}),D(I_{Ref},I_{Dist2}))),
\end{aligned}
\end{equation}
where $h\in(0,1)$ is the ground-truth judgment label. The total training loss is composed of two parts:
\begin{equation}
\label{equ:total}
L_{total} = L_{bce} + \lambda_{reg} L_{reg},
\end{equation}
where $\lambda_{reg}$ is the hyper parameter that balances the weight of the two loss items.

The final predicted results can be given by:
\begin{equation}
\label{eq:d1+d2}
{{h}^{*}}=\left\{ \begin{aligned}
  & 0, D(I_{Ref},I_{Dist1})<=D(I_{Ref},I_{Dist2}) \\ 
 & 1, D(I_{Ref},I_{Dist1})>D(I_{Ref},I_{Dist2}) \\ 
\end{aligned} \right.
\end{equation}
And the judgment accuracy can be calculated by:
\begin{equation}
    Acc =\tfrac{\sum\limits_{i=1}^{N}{(h_{i}^{*}=={{h}_{i}})}}{N},
\end{equation}
where $N$ is the total number of the triplet $(I_{Ref},I_{Dist1},I_{Dist2})$.
\section{Experiments}
\subsection{Datasets}
We summarize the datasets we use for pre-training, training and testing in Table \ref{tab:datasets}. During the pre-training stage, we only use KADID-10K\cite{KADID-10K} datsest for training.
Specially, CLIC datasets consist of images generated by various compression methods including traditional codecs(\textit{e.g.}, HEVC/H.265 \cite{hevc}, VVC/H.266 \cite{vvc}) and learning-based methods~\cite{wu2020learned,gao2021perceptual}. In order to cover the distortion types as comprehensively as possible, we select three another datasets: PIPAL\cite{pipal}, BAPPS\cite{LPIPS} and PieAPP\cite{pieapp}, which include both traditional distortions and algorithm outputs to join in the training process.  We split
109,896 triplets out of the CLIC2021Test for training (\textit{i.e.}, CLIC2021Test-subtrain) and the remaining 
12,211 triplets for testing (\textit{i.e.}, CLIC2021Test-subtest). We also use CLIC2022Val dataset for tesing.

\begin{table*}[htbp]
\centering
\caption{Summarization of datasets we use for pre-training, training and testing.}
\scalebox{0.9}{
\begin{tabular}{cc|cccc}
\hline
\multicolumn{2}{c|}{Dataset} & \begin{tabular}[c]{@{}c@{}}Num\\ Distort.\end{tabular} & \begin{tabular}[c]{@{}c@{}}Distort.\\ Types\end{tabular} & \begin{tabular}[c]{@{}c@{}}Distort.\\ Images/Patches\end{tabular} & \begin{tabular}[c]{@{}c@{}}Judgment\\ Type\end{tabular} \\ \hline
\multicolumn{1}{c|}{Pre-training} & KADID-10K\cite{KADID-10K} & 25 & traditional & 10.1k & MOS \\ \hline
\multicolumn{1}{c|}{\multirow{5}{*}{Training}} & PIPAL\cite{pipal} & 40 & trad.+alg.outputs & 29k & MOS(Elo system) \\
\multicolumn{1}{c|}{} & BAPPS(2AFC-Distort)\cite{LPIPS} & 425 & trad.+CNN & 321.6k & 2AFC \\
\multicolumn{1}{c|}{} & BAPPS(2AFC-Real alg)\cite{LPIPS} & - & alg.outputs & 53.8k & 2AFC \\
\multicolumn{1}{c|}{} & PieAPP\cite{pieapp} & 75 & trad.+alg.outputs & 20.3k & 2AFC \\ 
\multicolumn{1}{c|}{} & CLIC2021Test-subtrain & - & codec outputs & 109.9k & 2AFC \\ \hline
\multicolumn{1}{c|}{\multirow{2}{*}{Testing}}  & CLIC2021Test-subtest & - & codec outputs & 12.2k & 2AFC \\
\multicolumn{1}{c|}{}& CLIC2022Val & - & codec outputs & 5.2k & 2AFC \\\hline
\end{tabular}}
\label{tab:datasets}
\vspace{-2mm}
\end{table*}

\subsection{Implementation Details}
To balance the performance and the computational complexity, we adopted Swin-T as the backbone which consists of 4 stages (layer numbers={2, 2, 6, 2}). The linear embedding dimension $C$ of stage one was set to 96. The patch size was set to 4 and window size was set to 7.
SwinIQA was first pre-trained by optimizing the objective in Eq.~\ref{equ:reg}. We trained the network on KADID-10K for 50 epochs, with a batch size of 48 and a learning rate of $1e^{-4}$. 
The training of the SwinIQA was carried out by optimizing the objective in Eq.~\ref{equ:total} with the learning rate of $1e^{-4}$. The learning rate of the judgment network $\mathcal{G}$ in Eq.~\ref{equ:bce} was also set to $1e^{-4}$.  The value of $\lambda_{reg}$ was set to  5.0. 
We randomly cropped the images to $224 \times 224 \times 3$ while training. During testing, we cropped the images into various patches and averaged the predicted distances of all patches to get more accurate results.

\subsection{Discussion of different distance mapping strategies}
In this section, we discuss the performance of 5 different distance mapping strategies.
\begin{itemize}
    \item Mode 1: ${{f}_{mapped}}={{({{f}_{Dis}}-{{f}_{\operatorname{Re}f}})}^{2}} \circledast f_{Ref}$, where $\circledast$ denotes the cross attention opreration.
    \item Mode 2: $f_{mapped}=f_{Dist} \circledast f_{Ref}$.
    \item Mode 3: $f_{mapped}=(f_{Dist}-f_{Ref}) \circledast f_{Ref}$.
    \item Mode 4: $f_{mapped}=(f_{Dist}-f_{Ref})$.
    \item Mode 5: $f_{mapped}=f_{Dist} \circledcirc f_{Ref}$, where $\circledcirc$ means the similarity distance used in DISTS\cite{DISTS}.
\end{itemize}
The results on the KADID-10K testing set is shown in Table \ref{tab:distance mapping}. From the table, we can see that the cross attention between  ${{({{f}_{Dis}}-{{f}_{\operatorname{Re}f}})}^{2}}$ and $f_{Ref}$  is the most effective mapping strategy when comparing the perceptual similarity of two image representations.

\begin{table}[]
\centering
\caption{PLCC and SROCC performance on KADID-10K of different distance mapping strategies.}
\scalebox{0.9}{
\begin{tabular}{|c|c|c|}
\hline
Mode & PLCC & SROCC \\ \hline
1 & \textbf{0.9521} & \textbf{0.9553} \\ \hline
2 & 0.9213 & 0.9270 \\ \hline
3 & 0.9451 & 0.9482 \\ \hline
4 & 0.8698 & 0.8718 \\ \hline
5 & 0.7713 & 0.7311  \\ \hline
\end{tabular}}
\label{tab:distance mapping}
\vspace{-3mm}
\end{table}

\subsection{Comparisons with state-of-the-arts}
We compare our method with two traditional methods (PSNR and MS-SSIM), two CNN-based methods (LPIPS and DISTS), one Transformer-based method IQT\cite{cheon2021perceptual} and last year's champion method MMFN\cite{peng2021multi}. All the compared learning-based methods are retrained using the same datasets as SwinIQA. Given triplets $(I_{Ref},I_{Dist1},I_{Dist2})$, we record the predicted judgment (which distorted image is closer to the reference image $I_{Ref}$) given by each metric and compute the accuracy. We evaluate our performance on CLIC2022 validation set (CLIC2022Val) and the subtest of CLIC2021 testing set (CLIC2021Test-subtest) . The comparison results can be found in Table \ref{Tab:expe_results}. Our method steadily outperforms other methods regarding the compressed images.
\vspace{-2mm}
\begin{table}[htbp]
\centering
\caption{Accuracy evaluation on CLIC2022Val and CLIC2021Test-subtest.}
\scalebox{0.9}{
\begin{tabular}{c|cc}
\toprule[1.3pt]
Methods  & \begin{tabular}[c]{@{}c@{}}CLIC2022Val\\\end{tabular} & \begin{tabular}[c]{@{}c@{}}CLIC2021Test-\\ subtest\end{tabular} 
\\ \hline
PSNR     &0.572 &  0.510   \\
MS-SSIM\cite{ms-ssim}  &0.612 & 0.525 \\
LPIPS\cite{LPIPS}   &0.761 & 0.749   \\
DISTS\cite{DISTS}   &0.762& 0.752  \\ 
IQT\cite{cheon2021perceptual}    &0.766& 0.767    \\
MMFN \cite{peng2021multi}    &0.764& 0.753  \\\hline
Ours    &\textbf{0.780}  &\textbf{0.773}   \\ \hline                   
\end{tabular}
}
\label{Tab:expe_results}
\vspace{-4mm}
\end{table}


\section{Conclusion}
In this paper, we propose a full-reference image quality metric SwinIQA for compressed images. We employ Swin Transformer to extract the hierarchical feature representa-
tions. Then we utilize the cross attention operation to map the pair of reference and distorted image representations to the learned Swin distance space. Extensive experiments have demonstrated the effectiveness of the proposed SwinIQA for the perceptual quality assessment of compressed images.

\section*{Acknowledgement}
This work was supported in part by NSFC under Grant U1908209, 62021001 and the National Key Research and Development Program of China 2018AAA0101400.

{\small
\bibliographystyle{ieee_fullname}
\bibliography{egbib}
}

\end{document}